\documentclass[a4paper, 10pt, conference]{ieeeconf}      % Use this line for a4
\IEEEoverridecommandlockouts                              % This command is only
\usepackage[utf8]{inputenc}
\usepackage[T1]{fontenc}
\usepackage[table,xcdraw]{xcolor}
\usepackage{graphics} % for pdf, bitmapped graphics files
\usepackage{epsfig} % for postscript graphics files
\usepackage{hyperref}
\usepackage{multirow}
\usepackage{booktabs}
\usepackage{subfigure}
\usepackage{comment}

\graphicspath{ {images/} }
\title{\LARGE \bf
Scene Understanding for Autonomous Driving
}

\author{{
\parbox{7 in}{
    \centering
    \`Oscar Lorente Corominas \qquad
    Ian Riera Smolinska \qquad
    Aditya Sangram Singh Rana \\
    Master in Computer Vision\\
    Universitat Aut\`onoma de Barcelona\\
    08193 Bellaterra, Barcelona, Spain\\
    {\tt\small\{oscar.lorentec, 
        ianpau.riera,
        adityasangramsingh.rana\}@e-campus.uab.cat
    }}
}}

\begin{document}

\maketitle
\thispagestyle{empty}
\pagestyle{empty}

\begin{abstract}
    To detect and segment objects in images based on their content is one of the most active topics in the field of computer vision. Nowadays, this problem can be addressed using Deep Learning architectures such as Faster R-CNN or YOLO, among others. In this paper, we study the behaviour of different configurations of RetinaNet, Faster R-CNN and Mask R-CNN presented in Detectron2. First, we evaluate qualitatively and quantitatively (AP) the performance of the pre-trained models on KITTI-MOTS and MOTSChallenge datasets. We observe a significant improvement in performance after fine-tuning these models on the datasets of interest and optimizing hyperparameters. Finally, we run inference in unusual situations using out of context datasets, and present interesting results that help us understanding better the networks.
\end{abstract}

\section{Introduction}
During the last years, there has been a great increase in the number of applications in which object detection and instance segmentation are useful. Helping people organise their photo collections, analysing medical images or identifying what's around self-driving cars are just a few examples. These tasks require precisely labeled large-scale datasets, and most of them include a huge variety of object classes, from dogs or cats, to cars, boats, and so on.

\begin{figure}[b!]
    \centering
    \includegraphics[width=0.7\linewidth]{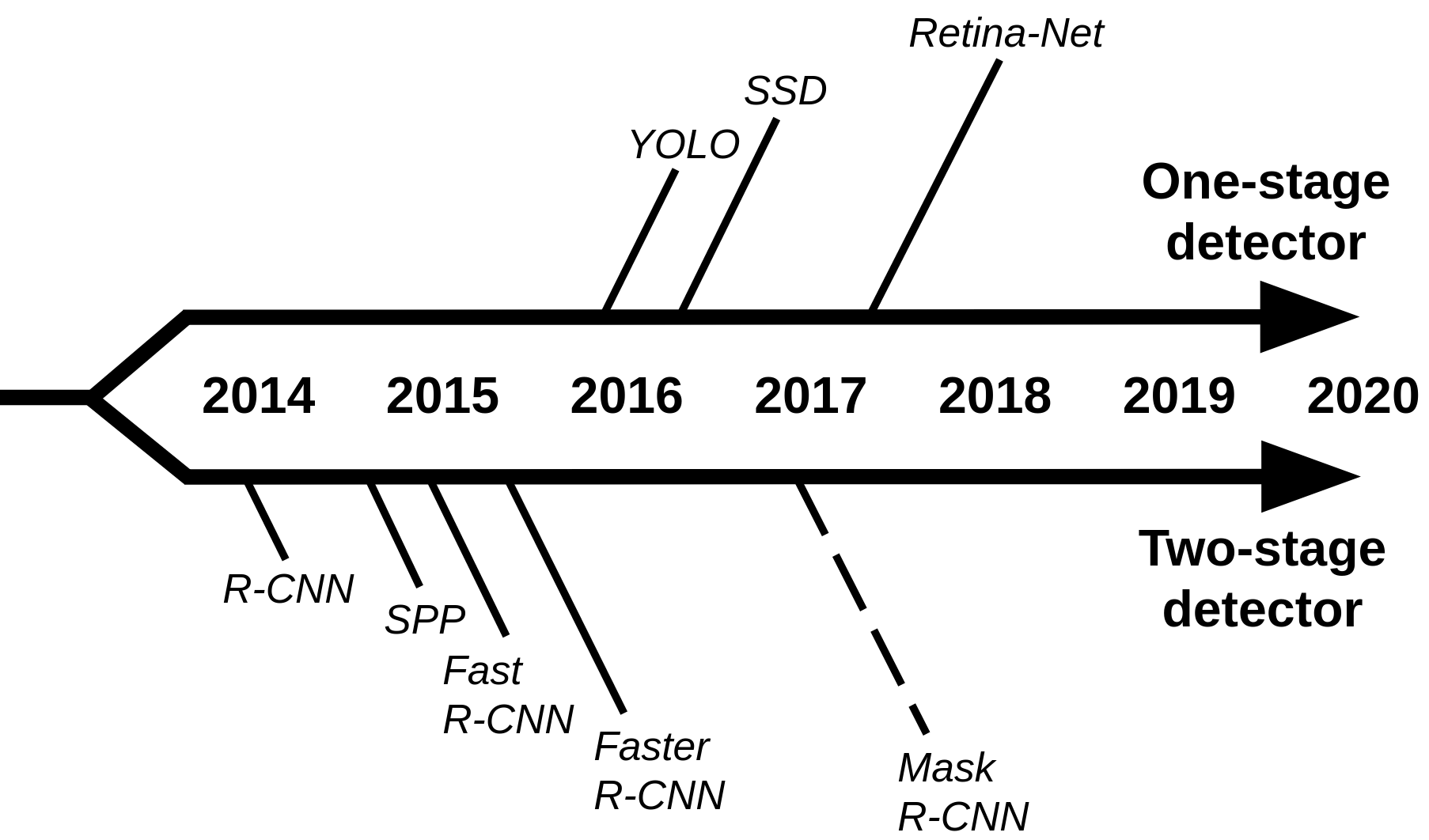}
    \caption{Simplified chronology of state-of-the-art object detection and instance segmentation architectures.}
    \label{fig:sota_chronology}
\end{figure}

In object detection and segmentation, given an input image, the goal is to locate objects and predict the class they belong to, and even segment them at a pixel level. This is not a big deal for humans, but teaching computers to \textit{see} is a difficult problem that has become a broad area of research interest. Advances in technology allow the use of Deep Learning techniques to automatically detect objects in scenes, and many different architectures have been developed in the past few years (Fig.~\ref{fig:sota_chronology}).

In this report, we present a deep study of three modern networks: RetinaNet~\cite{retina}, Faster R-CNN~\cite{faster_rcnn} and Mask R-CNN~\cite{mask_rcnn}. Different configurations from Detectron2~\cite{detectron2} are evaluated both qualitatively and quantitatively (using AP) on two datasets based on KITTI~\cite{kitti}. As a further research, we explore the behaviour of these architectures in unusual (and probably unseen) situations, using out of context datasets. This way, we try to better understand what are the networks actually learning.

\section{Related Work}

\begin{figure}[t!]
    \centering
    \includegraphics[width=0.8\linewidth]{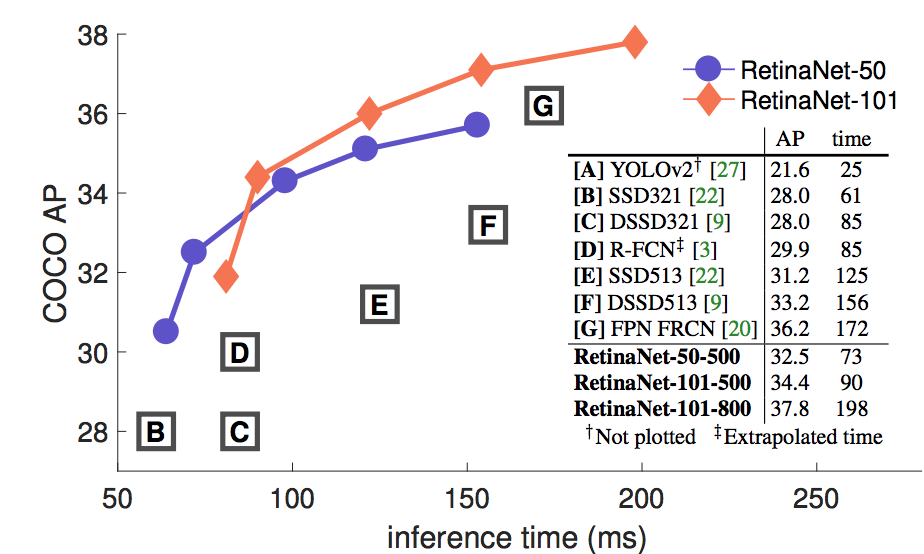}
    \caption{Comparison of AP on COCO dataset vs inference time (ms) between object detection models.} 
    \label{fig:comparison}
\end{figure}

\subsection{Two-stage detectors}

\textbf{R-CNN} \cite{rcnn} uses selective search \cite{selective_search} to extract 2000 regions from an image, which are called Regions of Interest (RoI). The RoI are then reshaped so that they match the input size of a pre-trained CNN, that extracts features for each region and a Support Vector Machine (SVM) is used to classify them. Finally, a linear regression model is trained to generate bounding boxes for each identified object.

R-CNN takes around 47 seconds for each image (inference), and the training stage is expensive and slow, as it extracts features with a CNN from 2000 regions per image.

A possible solution to this problem comes from \textbf{Spatial Pyramid Pooling} \cite{spp} (\textbf{SPP}), which goal is to get fixed-length representations for variable size feature maps. In this case, a CNN is run just once per image to obtain a feature map, and then use a (variable size) window related to the region proposals to detect the objects in the image.

This idea is extended in \textbf{Fast R-CNN} \cite{fast_rcnn}, which uses a method called RoI pooling similar to SPP. The RoI pooling layer reshapes the region proposals to the size of the CNN input. Finally, each region is passed on to a Fully Connected Network (FCN), where a softmax layer and a linear regression layer output the classes and bounding box coordinates. In Fast R-CNN, a joint loss is used.

Fast R-CNN solves two major issues of R-CNN: it passes one instead of 2000 regions per image to a CNN, and it uses one instead of three different models for extracting features, classification and generating bounding boxes. However, now the bottleneck is in the selective search algorithm used to find the RoI, which is slow and time consuming, leading to a final inference time of around 2 seconds per image.

To solve this problem, \textbf{Faster R-CNN} \cite{faster_rcnn} introduces Region Proposal Networks (RPN): using a sliding window over the feature maps from a CNN, it generates 9 anchor boxes of different shapes and sizes at each sliding position. For each anchor, RPN predicts two things: the probability that an anchor is an object (class-agnostic), and the bounding box regressor for adjusting the anchors to better fit the object.

In Faster R-CNN, the inference time is around 0.2 seconds per image, which is much faster than R-CNN and Fast R-CNN, but is still far from real-time object detection. 

\subsection{Object instance segmentation}
Towards segmentation, Mask R-CNN \cite{mask_rcnn} extends Faster R-CNN by adding a branch for predicting an object mask in parallel with the existing branch for bounding box recognition. It applies segmentation over the RPN predictions, generating a high-quality segmentation mask for each instance. 

\subsection{One-stage detectors}

Instead of using region proposals, one-stage detectors make predictions by only looking into the input image once, which can result in a faster performance. One of the state-of-the-art single pass detectors is \textbf{YOLO} \cite{yolo} (and its successors \cite{yolov2, yolov3, yolov4}), which idea is to extract a feature map using a CNN called Darknet, divide it into $SxS$ cells (YOLO uses $S=7$), and predict one bounding box for each cell. YOLO predicts the class of that bounding box and if it is centered at that cell, and uses Non-Maximum Suppression (NMS) to reduce the number of output bounding boxes.

Another well-known one stage detector is \textbf{SSD} \cite{ssd}, which makes predictions at multiple feature maps (YOLO only does it for one) using a VGG16 network \cite{vgg} as a feature extractor. The idea is that each feature map, which has different local receptive fields, specializes in objects of different sizes.

The NMS method is also used at the end of the SSD model, and Hard Negative Mining (HNM) is then used to further reduce the number of predicted negative boxes.

Finally, \textbf{RetinaNet} \cite{retina} has been formed by making two improvements over existing single stage detectors: Feature Pyramid Networks (FPN) \cite{fpn} and Focal Loss. FPN replaces the feature extractor of detectors like Faster R-CNN, and focal loss is introduced to handle the class imbalance problem with one-stage object detection models.

A comparison between some one-stage and two-stage detectors on COCO dataset \cite{coco} is presented in Fig.~\ref{fig:comparison} (source: \cite{retina}). As observed, Faster R-CNN ([G]) provides high Average Precision (AP) values, but the inference time is higher. On the other hand, the one-stage detectors are much faster, being YOLO the fastest one, but the AP is lower in most cases. There is a trade-off between computational speed and accuracy. In some cases, one-stage detectors can be more precise than two-stage, such as RetinaNet-101, but with a high computational cost as well.

\begin{figure}[b!]
\centering
\includegraphics[height=2.5cm]{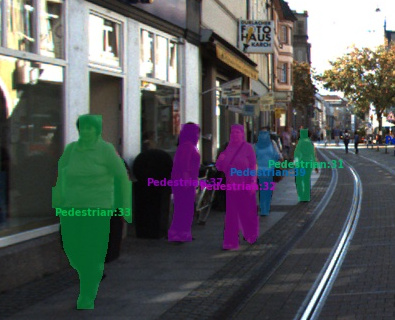}
\includegraphics[height=2.5cm]{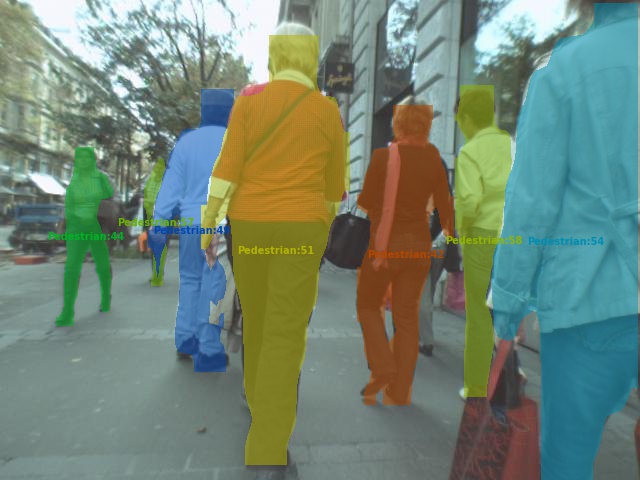}
\caption{\label{fig:kitti-and-mot-samples} Sample Annotations from the Datasets: KITTI MOTS (left) and MOTSChallenge (right).}
\end{figure}

\section{Experimentation setup}

\subsection{Datasets}

We present experimental results on the detection track of the 5th BMTT MOTChallenge Workshop: Multi-Object Tracking and Segmentation. The datasets used comprise of the KITTI MOTS and the MOTSChallenge datasets as presented in MOTS: Multi-Object Tracking and Segmentation~\cite{kittimots}. They are based on the KITTI Vision Benchmark Suite~\cite{kitti} and Multiple Object Tracking Benchmark~\cite{mot}. Sample annotations are presented in Fig. \ref{fig:kitti-and-mot-samples}.

\subsection{Train-Val-Test Split}
\label{sec:cross-val}
We use the official validation set of KITTI MOTS as our test set and the training sets of KITTI MOTS (12 sequences) and MOTSChallenge (4 sequences) for training and validation. The detailed statistics are provided in Appendix~\ref{sec:app_dataset}. 

We perform 4-fold cross validation with each fold containing 3 sequences from KITTI MOTS and 1 sequence from MOTSChallenge. For each experiment, we train with 3 folds and validate the remaining one for all 4 possible combinations. 

\subsection{Evaluation Metrics}
We compare the performance of the object detection and instance segmentation models using Average Precision (AP), after setting a threshold for the Intersection over Union (IoU). 

The general AP value defined for the COCO dataset consists of averaging all the AP for IoU from 0.5 to 0.95 with a step size of 0.05. Furthermore, in this case, the evaluation is presented in three different metrics that represent the size of the objects: small (APs), medium (APm) and large (APl). 

\subsection{Object Detection and Instance Segmentation Models}

We use Detectron2 \cite{detectron2}, an object detection platform of Facebook AI Research (FAIR) project. It includes a model zoo of state-of-the-art object detection and instance segmentation algorithms such as Faster R-CNN, RetinaNet and Mask R-CNN. Each model configuration has four parts:
\begin{itemize}
    \item Backbone: ResNet (R) or ResNext (X)
    \item Number of layers: 50 or 101
    \item Backbone combination: ResNet + Feature Pyramid Network (FPN), ResNet conv4 backbone with conv5 head (C4) or ResNet conv5 with dilations (DC5)
    \item Learning Rate (LR) Scheduler: 1x or 3x
\end{itemize}

For example, Mask R-CNN R\_50\_FPN\_3x uses ResNet with 50 layers as backbone, a FPN and a 3x LR scheduler.

\section{Results}
\subsection{Object Detection}
\subsubsection{Inference with pre-trained models}

\begin{table}[t!]
\centering
\begin{tabular}{ccc}
\toprule
 & \multicolumn{2}{c}{{Model}} \\ 
Metric & RetinaNet & Faster R-CNN \\ 
\midrule
AP   & 56.63 & 55.61 \\
APs  & 29.92 & 30.21 \\
APm  & 62.75 & 61.20 \\
APl  & 73.13 & 70.40 \\
\bottomrule
\end{tabular}
\caption{Pre-trained RetinaNet and Faster R-CNN R\_50\_FPN\_3x models evaluated on the KITTI MOTS validation dataset}
\label{tab:detection_initial_comparison}
\end{table}

We try out different configurations of Faster-RCNN and RetinaNet available in Detectron 2 pre-trained with COCO dataset. We decide to study the behaviour of each model qualitatively, and then evaluate their performance quantitatively to choose the best candidate for fine tuning it on our datasets. See Appendix \ref{sec:app_det_qualitative_quantitative} for the details of this experiments.

The best results in both cases are obtained with the configuration \textit{R-50-FPN-3x}, so it is the one used for fine tuning and hyperparameter optimization. We summarize the results in Table \ref{tab:detection_initial_comparison}. In terms of size of the bounding boxes, the best results are obtained using the APl metric, so both models are able to detect objects with a higher precision if they are larger (cars). As expected, the smaller the objects, the lower the precision (APs). 

\begin{table}[b!]
\centering
\begin{tabular}{ccccccc}
\toprule
Architecture & \multicolumn{2}{c}{{FasterRCNN}} & \hspace{0.1cm} & \multicolumn{2}{c}{RetinaNet}\\ 
 \cmidrule{2-3} 
 \cmidrule{5-6} 
AP & Pretrained & Finetuned && Pretrained & Finetuned \\ 
\midrule
Mean & 55.61 & 60.31 && 56.63 & 61.47 \\
Car & 65.72 & 66.93 && 67.14 & 67.84 \\
Pedestrian & 45.50 & 53.69 && 46.13 & 55.1 \\
\bottomrule
\end{tabular}
\caption{Comparison of results before and after fine-tuning the models on the KITTI MOTS and MOTSChallenge datasets}
\label{tab:after_finetune}
\end{table}

\subsubsection{Hyperparameter Optimization}
To improve the results, we train and fine tune the model that gave us the best inference results. We run a limited non recursive search using cross-validation: first on LR (1e-3, 1e-4 and 1e-5), and then on the batch size - number of ROI per image (256, 512 and 1024). The best performance is obtained with a LR of 1e-4 and a batch size of 256 for Faster R-CNN, and 1e-3 and 512 for RetinaNet, respectively. The complete results can be found in Appendix~\ref{sec:app_det_hyperparameter}.

\subsubsection{Evaluation on test data}

After optimizing the hyperparameters of each model using cross-validation, we train them using the entire training data available and test them against the 9 sequences of KITTI-MOTS official validation set.

\begin{figure}[t!]
    \centering
    \includegraphics[width=0.75\linewidth]{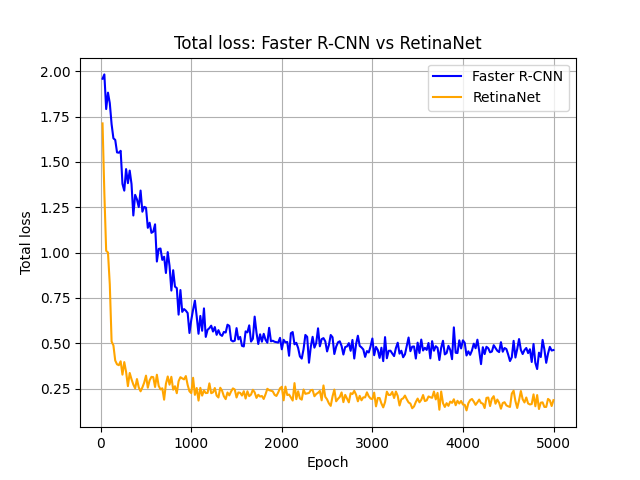} 
    \caption{Total loss of Fast-RCNN vs RetinaNet.} \label{fig:final_total_loss}
\end{figure}

The total loss obtained with the optimized Faster R-CNN and RetinaNet models is presented in Fig.~\ref{fig:final_total_loss}. As observed, RetinaNet converges faster to a lower minimum loss.

The results of both models before and after fine tuning are presented in Table \ref{tab:after_finetune}. The performance is boosted after fine tuning the models to KITTI-MOTS dataset, as it learns to detect objects in those scenarios more precisely and thus the resulting AP is better. RetinaNet performs better than Faster R-CNN, and in both cases the improvement is found in the pedestrian class. The performance in the car class is already acceptable even without fine tuning, as the bounding boxes are bigger and easier to detect.

%% ------------------------ Instance segmentation ------------------------

\subsection{Object Instance Segmentation}
\subsubsection{Inference with pre-trained models}

We try out different configurations of Mask-RCNN available in Detectron 2. All the models have been pre-trained with COCO, except one that has been also fine tuned with the Cityscapes dataset~\cite{cityscapes} (R\_50\_FPN\_3x). First, we evaluate qualitatively and quantitatively the models pre-trained with COCO. See Appendix \ref{sec:app_mask_qualitative_quantitative}  for the details of this experiments.

The best results are obtained with the configuration \textit{R-50-FPN-3x}, which is the one that has another version fine-tuned with Cityscapes. The comparison between both models is presented in Table \ref{tab:coco_cityscapes_initial}. In terms of instance segmentation (Mask AP), we can observe a correlation with the detection results (Box AP), but with a general downgrade on the obtained accuracy. This makes sense, as segmenting the shape of the objects with precision at pixel level is a harder task than just bounding boxing the object. The downgrade is especially notable on small and medium objects (APs and APm), as it is harder to segment them than larger objects. This also explains the difference in performance of the model on the car and pedestrian classes.

\begin{table}[t!]
\centering
\begin{tabular}{ccc|cc}
\toprule
& \multicolumn{2}{c}{{COCO}} & \multicolumn{2}{c}{{Cityscapes}} \\ 
Metric & Box & Mask & Box & Mask \\ 
\midrule
AP   & 56.96 & 45.90 & 51.11 & 41.45 \\
APs  & 40.94 & 27.93 & 36.33 & 26.45 \\
APm  & 67.91 & 55.79 & 60.93 & 50.83 \\
APl  & 64.11 & 65.18 & 64.76 & 57.11 \\
\midrule
Car & 67.45 & 60.74 & 58.41 & 54.57 \\
Pedestrian & 46.46 & 31.05 & 43.80 & 28.43 \\
\bottomrule
\end{tabular}
\caption{Mask R-CNN R\_50\_FPN\_3x model pre-trained on COCO and Cityscapes: evaluation on the KITTI MOTS validation dataset}
\label{tab:coco_cityscapes_initial}
\end{table}

In general terms, both box and mask results are worse when using the model trained with Cityscapes. The worse results on the class pedestrian could be explained by the fact that Cityscapes distinguishes between persons on bicycles (rider class) or persons on foot, while our dataset ground truth considers all of them pedestrians.

\subsubsection{Training on different datasets}
After evaluating the pre-trained models, we train the R\_50\_FPN\_3x Mask RCNN model on KITTI-MOTS and MOTSChallenge training sets and evaluate it on KITTI-MOTS official validation set. The goal is to compare the performance of the model before and after training it on different combinations of datasets. See Appendix~\ref{sec:app_mask_train_datasets} for a detailed analysis of the qualitative results.

\begin{table}[b!]
\centering
\begin{tabular}{ccc|cc|cc}
\toprule
 & \multicolumn{2}{c}{\multirow{2}{*}{{COCO}}} & \multicolumn{2}{c}{\multirow{2}{*}{{+ KITTI-MOTS}}} & \multicolumn{2}{c}{{+ KITTI-MOTS}} \\
{} & {} & {} & {} & {} & \multicolumn{2}{c}{+ MOTSChallenge} \\
\midrule
 & Box & Mask & Box & Mask & Box & Mask\\ 
\midrule
Mean & 59.96 & 45.90 & 59.37 & 48.83 & 59.47 & 46.39 \\
Car & 67.45 & 60.74 & 68.74 & 63.79 & 65.56 & 61.79 \\
Pedestrian & 46.46 & 31.05 & 49.99 & 33.86 & 53.37 & 30.99\\
\bottomrule
\end{tabular}
\caption{Mask R-CNN R\_50\_FPN\_3x model pre-trained on COCO and fine-tuned on different datasets}
\label{tab:maskrcnn_coco}
\end{table}

In Table~\ref{tab:maskrcnn_coco} we present a quantitative comparison of the model trained on COCO, with and without the KITTI-MOTS and MOTSChallenge datasets. We can observe an improvement of 3\% when using the trained model with respect to the direct inference obtained using the pre-trained model on COCO. However, there is not much difference when adding MOTSChallenge dataset in terms of total AP. This is due to the fact that for class pedestrian the results improve, but they become worse for the class car.

\begin{table}[t!]
\centering
\begin{tabular}{ccc|cc|cc}
\toprule
 & \multicolumn{2}{c}{\multirow{2}{*}{{Cityscapes}}} & \multicolumn{2}{c}{\multirow{2}{*}{{+ KITTI-MOTS}}} & \multicolumn{2}{c}{{+ KITTI-MOTS}} \\
{} & {} & {} & {} & {} & \multicolumn{2}{c}{+ MOTSChallenge} \\
\midrule
 & Box & Mask & Box & Mask & Box & Mask\\ 
\midrule
Mean & 51.11 & 41.45 & \textbf{64.88} & \textbf{52.66} & 64.86 & 51.93 \\
Car & 58.41 & 54.57 & \textbf{71.79} & \textbf{66.39} & 70.20 & 64.37 \\
Pedestrian & 43.80 & 28.43 & 57.96 & 38.93 & \textbf{59.93} & \textbf{39.49}  \\
\bottomrule
\end{tabular}
\caption{Mask R-CNN R\_50\_FPN\_3x model pre-trained on COCO and Cityscapes and fine-tuned on different datasets}
\label{tab:maskrcnn_cityscapes}
\end{table}

When including the Cityscapes dataset (Table~\ref{tab:maskrcnn_cityscapes}), the results improve notably compared both to the direct inference and training with the COCO models. Again, using the MOTSChallenge dataset for training does not seem to improve the performance of the model, as it slightly improves the pedestrian class AP but worsens the class car AP. 

\textbf{The best results are obtained with the model pre-trained on COCO + Cityscapes, and then trained with KITTI-MOTS}. For this reason, this will be the combination of datasets used for the hyperparameter optimization.

\subsubsection{Hyperparameter Optimization}

We train and fine tune the model that gave us the best inference results using cross validation. The first hyper-parameter that we optimize is the LR scheduler. We run experiments without using any scheduler and using One-Cycle policy \cite{one_cycle}. The results in Table \ref{tab:oc_scheduler} show that using One Cycle Scheduler improve the performance in 60\% less iterations, so it is the one we will use from now on.

\begin{table}[b!]
\centering
\begin{tabular}{cc|cccc|c}
\toprule
Scheduler & \# iters & Box AP & Mask AP \\
\midrule
None & 5,000 & 57.57 & 46.44 \\
One Cycle & 2,000 & 60.54 & 46.80\\
\bottomrule
\end{tabular}
\caption{Box and Mask AP with and without One Cycle Scheduler}
\label{tab:oc_scheduler}
\end{table}

We run a limited non recursive search using cross-validation: first on the number of stages of ResNet to freeze (1 to 5), then on the background (bg) and foreground (fg) IOU thresholds and finally on the NMS threshold. The best performance is obtained freezing 2 stages, using 0.3 and 0.7 as the bg and fg thresholds and a NMS threshold of 0.7. The corresponding results for the different parameter optimizations can be seen in Appendix~\ref{sec:app_mask_hyperparameter}.

\subsubsection{Evaluation on test data}

After optimizing the hyperparameters of R\_50\_FPN\_3x, we train the final model using the entire training data available and test it against the 9 sequences of KITTI-MOTS official validation set.

The results of both models before and after fine tuning are presented in Table \ref{tab:final_results}. The performance is slightly improved after fine tuning the models to KITTI-MOTS dataset, as it learns to detect objects in those scenarios more precisely. However, it does not improve significantly, as they were already pre-trained with cars and pedestrians with two larger datasets (COCO+Cityscapes), and thus the weights were already trained to detect the classes of interest.

\begin{table}[t!]
\centering
\begin{tabular}{ccc|cc}
\toprule
 & \multicolumn{2}{c}{Pre-trained} & \multicolumn{2}{c}{Fine-tuned} \\
 & Box & Mask & Box & Mask\\ 
\midrule
Mean & 64.88 & 52.66 & 65.44 & 52.28 \\
Car & 71.79 & 66.39 & 72.25 & 66.23 \\
Pedestrian & 57.96 & 38.93 & 58.63 & 38.28 \\
\bottomrule
\end{tabular}
\caption{Mask R-CNN R\_50\_FPN\_3x model pre-trained on COCO and fine-tuned with the optimized hyperparameters}
\label{tab:final_results}
\end{table}

\section{Further Research}
Until now, all the results were presented using datasets with common objects in context, but we want to study the behaviour of the networks in unusual situations using \textbf{Out Of Context (OOC) datasets}. With this purpose, we use a small subset of images with OOC objects from SUN dataset\cite{sun09}, and we modify some specific images to perform experiments.

\begin{figure}[b!]
\centering
\begin{tabular}{cc}
\includegraphics[width=0.20\textwidth]{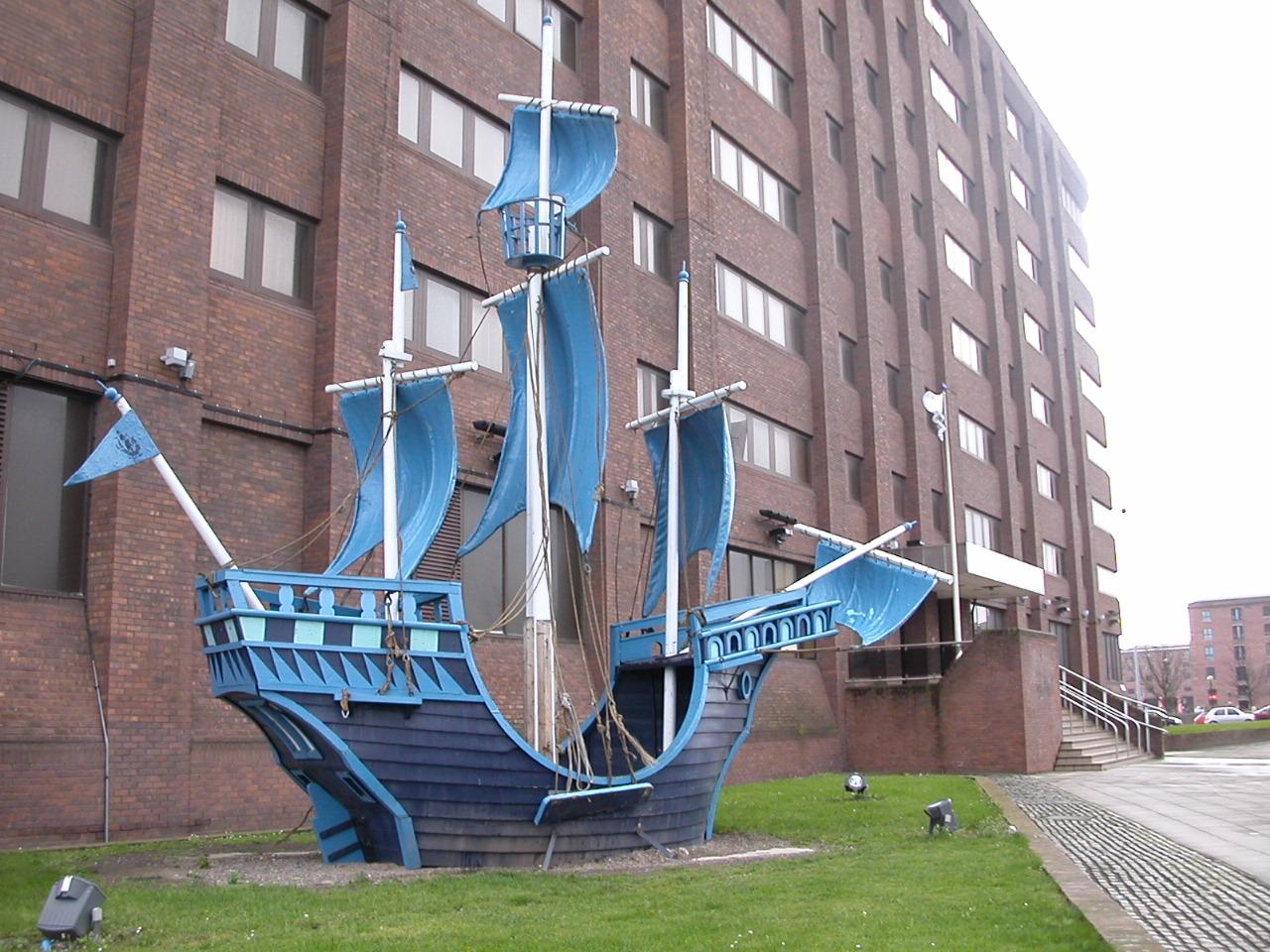} &
\includegraphics[width=0.20\textwidth]{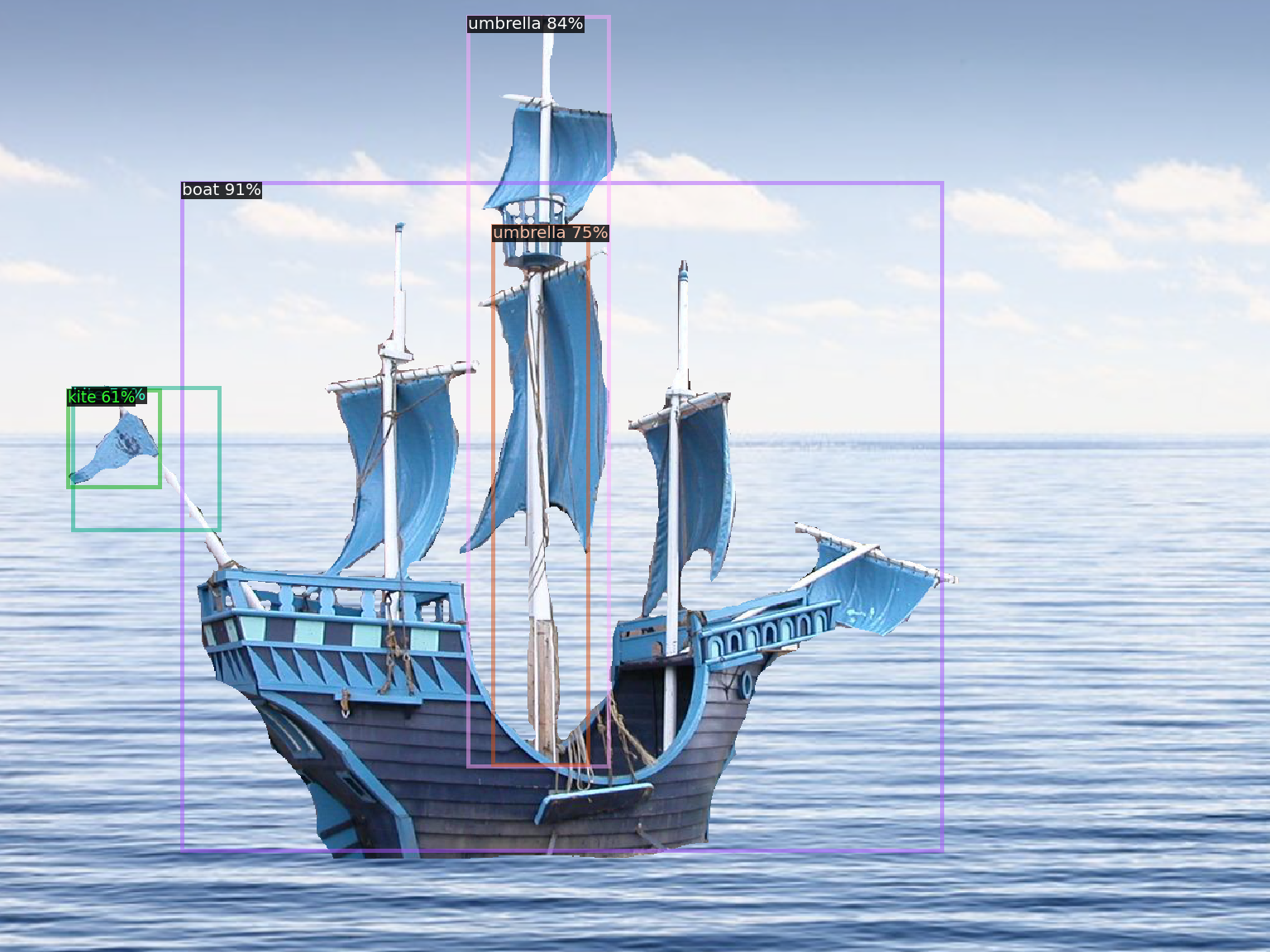} \\
(a) & (b) \\
\end{tabular}
\caption{\label{fig:boat} Detection of ship out of context (a) and in context (b).}
\end{figure}

In Fig.~\ref{fig:boat}a, we present a case of a ship not being detected, as it is not located where a ship is supposed to be: on the water. We validate this hypothesis in Fig.~\ref{fig:boat}b: now that the ship is in the sea, the model is able to recognize it, so the context is indeed relevant in this situation.

\begin{figure}[b!]
\centering
\begin{tabular}{cc}
\includegraphics[width=0.20\textwidth]{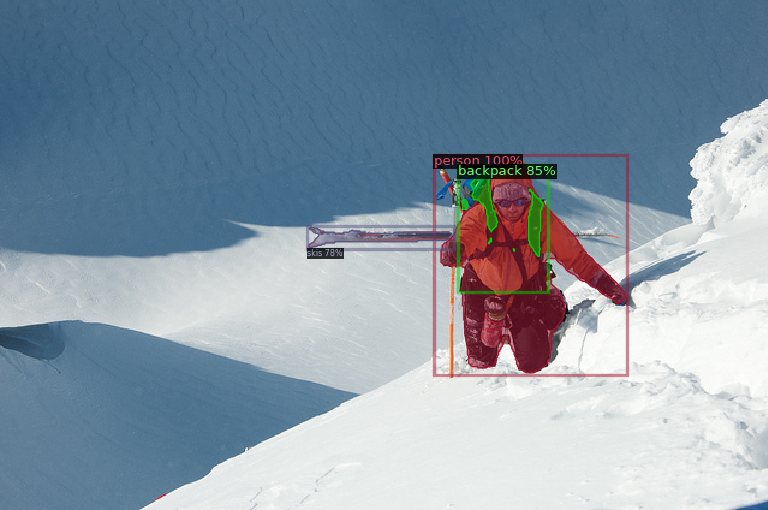} &
\includegraphics[width=0.20\textwidth]{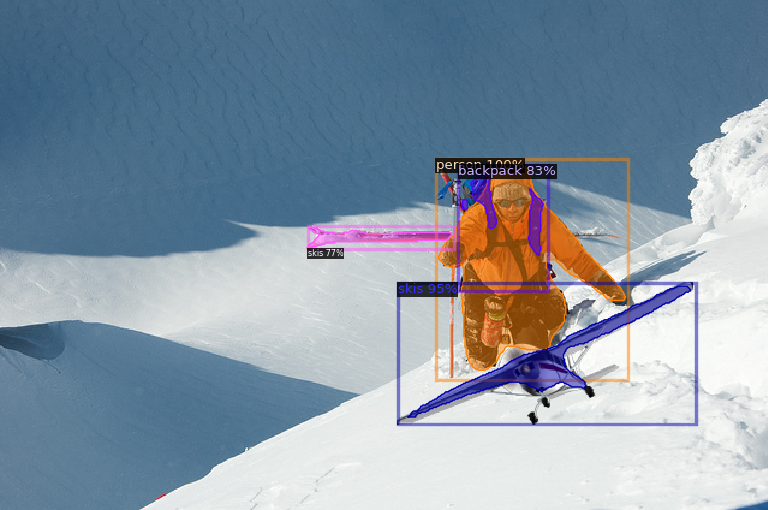} \\
(a) & (b) \\
\end{tabular}
\caption{\label{fig:ski} Segmentation of the original (a) and modified (b) images.}
\end{figure}

Another aspect to take into account when understanding scenes is the co-occurrence of objects. This case is studied in Fig.~\ref{fig:ski}, where a plane (that we manually added into the image) is detected as a ski when it's located below a person feet. It seems that the network takes into account the relative position between objects.

\begin{figure}[t!]
\centering
\begin{tabular}{cc}
\includegraphics[width=0.20\textwidth]{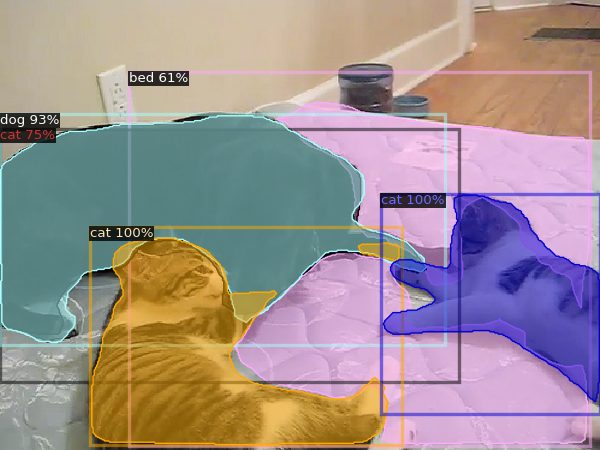} &
\includegraphics[width=0.20\textwidth]{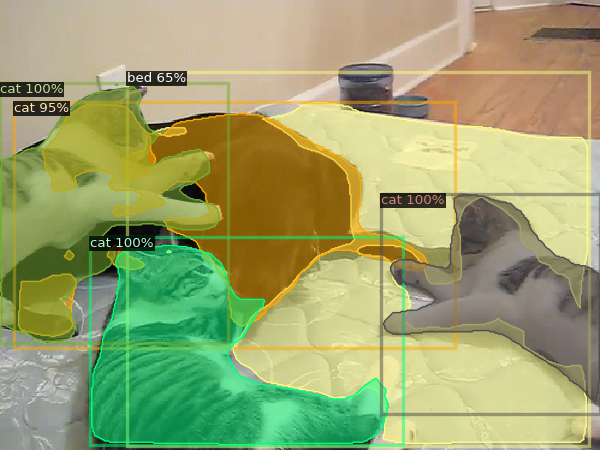} \\
(a) & (b) \\
\end{tabular}
\caption{\label{fig:cats} Segmentation of the original (a) and modified (b) images.}
\end{figure}

In Fig.~\ref{fig:cats} we duplicate a cat from within the image and copy it to another location in the same image. As observed, in the original image (Fig.~\ref{fig:cats}a) the dog is detected both as a dog and a cat (overlapped detection). When adding a cat (Fig.~\ref{fig:cats}b), the dog is now detected only as a cat, so it seems that the network relates the dog with the surrounding animals: if they are all cats, the dog has to be a cat as well.

\begin{figure}[t!]
\centering
\begin{tabular}{ccc}
\includegraphics[height=2.5cm]{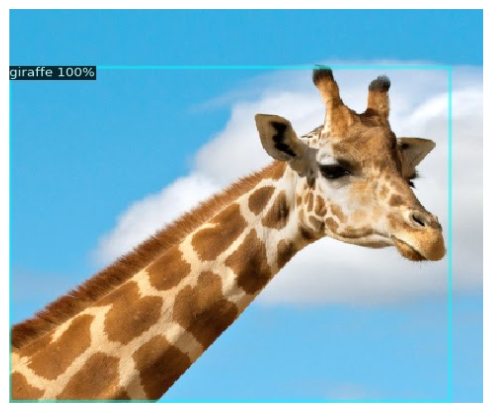} &
\includegraphics[height=2.5cm]{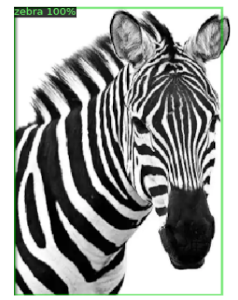} &
\includegraphics[height=2.5cm]{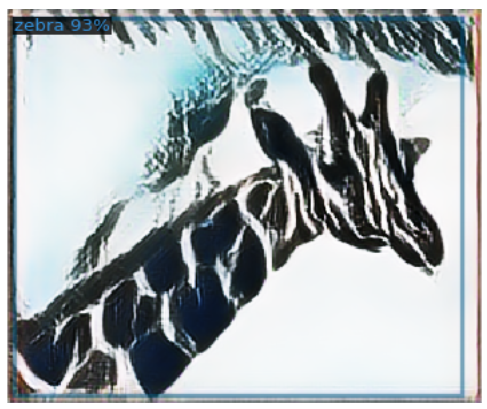} \\
(a) & (b) & (c) \\
\end{tabular}
\caption{\label{fig:style} Content from giraffe (a) and style from zebra (b) used to create a stylized image (c).}
\end{figure}

Finally, we use style transfer to explore the importance of texture and color to the detections. The results presented in Fig.~\ref{fig:style} show that both texture and color are important for the network, as it misclassifies the giraffe as a zebra when black and white strips are added to its body (Fig.~\ref{fig:style}c).

\section{Conclusions}
In this paper, we have analyzed the results obtained when applying object detection and segmentation on different scenarios. First, we tested pre-trained models to run inference on the KITTI-MOTS dataset, and we observed that as those models were pre-trained on large-scale datasets such as COCO, they generalized quite well to unseen scenes. For all the architectures tested, using FPN provided the best results when comparing different configurations of Detectron2.

Then, we fine-tuned the networks to our specific cases, training for the classes of interest and optimizing the hyper-parameters of the network. We observed an improvement on the performance when fine tuning with KITTI-MOTS dataset, but not much when including MOTSChallenge.

In general, AP results are better with larger objects (APm, APl), which are easier to detect than smaller ones (APs). This is also related to the AP of the car and pedestrian classes, the latter being the harder one to detect.

We observed a correlation between the obtained results on object detection and instance segmentation. However, the latter are slightly worse, as segmenting the shape of the objects with precision at pixel level is a harder task than just localizing the object in an image. This is especially notable with poor segmentation results on the pedestrian class, as it's harder to precisely segment smaller objects.

The discussed results were obtained with scenes where objects are in context. When running inference on images with objects that are out of context, the model struggled with unusual situations, providing worse qualitative results. Context, co-occurrence and texture, seem to affect the performance of the network when detecting and classifying objects.

We also showed that applying style transfer to the images drastically affects the model's ability to classify objects properly. The detection and segmentation ability is not affected as much as the network was still able to localize the objects in stylized images quite well.

\clearpage

\begin{appendices}
\section{Dataset}
\label{sec:app_dataset}
KITTI MOTS was obtained by annotating all the 21 sequences of KITTI tracking dataset~\cite{kitti} and MOTSChallenge was obtained  by annotating 4 of the 7 sequences of MOTChallenge2017 training dataset~\cite{mot}. KITTI MOTS contains annotations for both cars and pedestrians whereas MOTSChallenge focuses solely on pedestrians, especially in crowded scenes where tracking is very challenging due to many cases of occlusions. The statistics of the data used for training and testing can be found in Table \ref{tab:dataset-stats}.

\begin{table}[h]
\small
\setlength{\tabcolsep}{3.5pt}
\centering{}
    \begin{tabular}{lccc}
        \toprule 
         & \multicolumn{2}{c}{{\footnotesize{}KITTI MOTS}} & \multicolumn{1}{c}{{\footnotesize{}MOTSChallenge}}\tabularnewline
         & {\footnotesize{}train } & {\footnotesize{}test } & {\footnotesize{}train } \tabularnewline
        \midrule 
        {\footnotesize{}\# Sequences } & {\footnotesize{}12 } & {\footnotesize{}9 } & {\footnotesize{}4}\tabularnewline
        {\footnotesize{}\# Frames } & {\footnotesize{}5,027 } & {\footnotesize{}2,981 } & {\footnotesize{}2,862}\tabularnewline
        \midrule 
        {\footnotesize{}\# Masks Pedestrian } &  &  & \tabularnewline
        {\footnotesize{}\ \ \ \ \ \  Total } & {\footnotesize{}8,073 } & {\footnotesize{}3,347 } & {\footnotesize{}26,894 }\tabularnewline
        {\footnotesize{}\ \ \ \ \ \  Manually annotated } & {\footnotesize{}1,312 } & {\footnotesize{}647 } & {\footnotesize{}3,930 }\tabularnewline
        \midrule 
        {\footnotesize{}\# Masks Car } &  &  & \tabularnewline
        {\footnotesize{}\ \ \ \ \ \  Total } & {\footnotesize{}18,831 } & {\footnotesize{}8,068 } & {\footnotesize{}-}\tabularnewline
        {\footnotesize{}\ \ \ \ \ \  Manually annotated } & {\footnotesize{}1,509 } & {\footnotesize{}593 } & {\footnotesize{}-}\tabularnewline
        \bottomrule
    \end{tabular}
\caption{\label{tab:dataset-stats}Statistics of the modified split of KITTI MOTS and MOTSChallenge Datasets}
\end{table}

\section{AP and IOU}
\label{sec:app_ap_iou}
IoU is a metric that measures the overlap between 2 areas. In an object detection scenario, the IoU measures how much our predicted bounding box or mask overlaps with the ground truth. Hence, the IoU is computed for each prediction and a threshold is set to define whether the prediction is a true positive or a false positive. 
\begin{equation}
IoU = \frac{\textrm{area of overlap}}{\textrm{area of union}}  
\end{equation}

To calculate the AP, we first plot the precision-recall curve for a determined IoU threshold. The general definition of the AP is the area under this curve, but usually this operation is simplified. In our case, we use the simplification given by the COCO evaluator, consisting in sampling 101 points of the precision-recall curve after applying maximum precision to the right.

%----------------DETECTION
\section{Object Detection}

\subsection{Pre-trained models: Qualitative and Quantitative Results}
\label{sec:app_det_qualitative_quantitative}
\begin{figure}[t!]
    \centering
    \includegraphics[width=\linewidth]{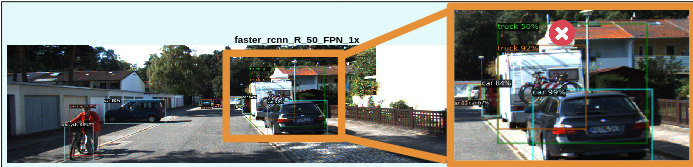}
    (a)
    \includegraphics[width=\linewidth]{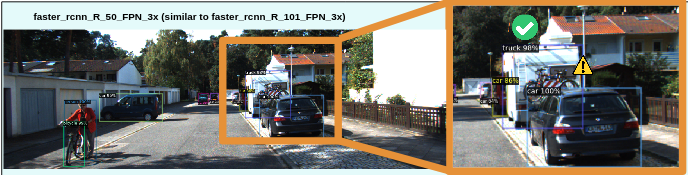}
    (b)
    \caption{Results from pre-trained Faster-RCNN models from Model Zoo: (a) R\_50\_FPN\_1x and (b) R\_50\_FPN\_3x}
    \label{fig:qualitative_faster}
\end{figure}

\begin{figure}[t!]
    \centering
    \includegraphics[width=\linewidth]{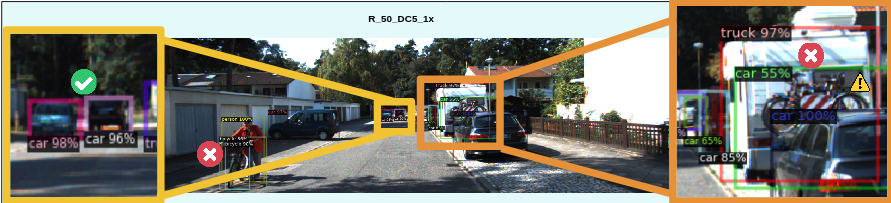}
    (a)
    \includegraphics[width=\linewidth]{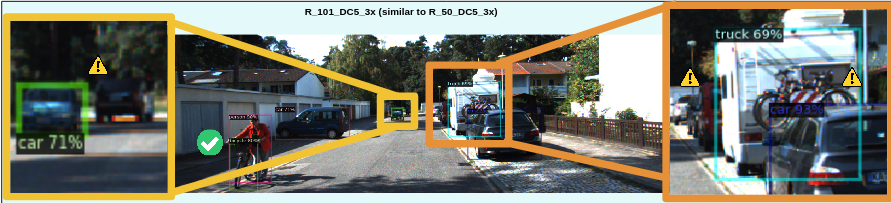}
    (b)
    \caption{Results from pre-trained Retina Net models from Model Zoo: (a) R\_50\_DC5\_1x and (b) R\_101\_DC5\_3x}
    \label{fig:qualitative_retina}
\end{figure}

For inference experiments, we try out different configurations of pretrained Faster-RCNN and RetinaNet architectures available in the Detectron 2 model zoo. In Figs.~\ref{fig:qualitative_faster} and ~\ref{fig:qualitative_retina} we present some qualitative results with different configurations for both models Faster R-CNN and RetinaNet. 

There is a false positive on the right part of the image for the R\_50\_FPN\_1x model (Fig.~\ref{fig:qualitative_faster}a), which is correctly detected when using R\_50\_FPN\_3x (Fig.~\ref{fig:qualitative_faster}b).

In Fig.~\ref{fig:qualitative_retina} we can observe that we obtain worse qualitative results with RetinaNet than Faster-RCNN. The detections in Fig.~\ref{fig:qualitative_retina}a shows the problem of double detection of the bicycle on the left and the detection of a ghost car on the right. On the other hand, Fig.~\ref{fig:qualitative_retina}b shows how the model fails to detect the car at the end of the street and the red van parked in front of the mobile home at the right. Additionally, none of the RetinaNet models detects the bicycles on the right (behind the mobile home).

\begin{table}[b!]
\centering
\begin{tabular}{cccc}
\toprule
 & \multicolumn{3}{c}{{Model: retinanet\_}} \\ 
\cmidrule{2-4} 
Metric & R50-FPN-1X & R50-FPN-3X & R101-FPN-3X \\ 
\midrule
AP   & 55.31 & \textbf{56.63} & 55.19 \\
APs  & 28.63 & \textbf{29.92} & 28.87 \\
APm  & 61.36 & \textbf{62.75} & 60.25 \\
APl  & 72.26 & \textbf{73.13} & 74.01 \\
\bottomrule
\end{tabular}
\caption{Pre-trained RetinaNet Models evaluated on the KITTI MOTS validation dataset}
\label{tab:retina_pre_trained}
\end{table}
\begin{table*}[t!]
\centering
\begin{tabular}{ccccccc|ccc}
\toprule
 & \multicolumn{9}{c}{{Model: FasterRCNN\_}} \\ 
\cmidrule{2-10}
& \multicolumn{6}{c|}{{R-50-X}} & \multicolumn{3}{c}{{R-101-X}}  \\
Metric & C4-1x & DC5-1x & FPN-1x & C4-3x & DC5-3x & FPN-3x & C4-3x & DC5-3x & FPN-3x \\ 
\midrule
AP   & 53.08 & 51.15 & 54.83 & 53.65 & 52.62 & \textbf{55.61} & 54.15 & 54.48 & 56.44 \\
APs  & 23.78 & 24.23 & 29.26 & 24.58 & 25.32 & \textbf{30.21} & 26.52 & 25.81 & 30.46 \\
APm  & 59.33 & 56.28 & 60.50 & 59.86 & 57.51 & \textbf{61.20} & 59.46 & 60.06 & 61.50 \\
APl  & 71.71 & 69.90 & 70.82 & 71.72 & 71.10 & \textbf{70.40} & 73.03 & 74.24 & 73.06 \\
\bottomrule
\end{tabular}
\caption{Pre-trained FasterRCNN Models evaluated on the KITTI MOTS validation dataset}
\label{tab:fasterrcnn_pre_trained}
\end{table*}

We summarize the quantitative results in Tables \ref{tab:retina_pre_trained} and \ref{tab:fasterrcnn_pre_trained}. The best results in both cases are obtained with the configuration \textit{R-50-FPN-3x}, so it is the one used for fine tuning and hyperparameter optimization.

\subsection{Hyperparameter Optimization}
\label{sec:app_det_hyperparameter}

\begin{table}[t!]
\centering
\begin{tabular}{c|cccc|c}
\toprule
LR & 0 & 1 & 2 & 3 & Mean AP \\
\midrule
1e-3 & 57.39 & 61.13 & 62.05 & 69.73 & 62.58 \\
1e-4 & \textbf{57.81} & \textbf{63.97} & \textbf{62.08} & \textbf{68.49} & \textbf{63.09} \\
1e-5 & 55.27 & 62.21 & 59.44 & 65.36 & 60.57 \\
\bottomrule
\end{tabular}
\caption{mAP for cross-val split for Model: faster\_rcnn\_R\_50\_FPN\_3x}
\label{tab:faster_hyper_lr}
\end{table}
\begin{table}[t!]
\centering
\begin{tabular}{c|cccc|c}
\toprule
Batch(ROI) & 0 & 1 & 2 & 3 & Mean AP \\
\midrule
\textbf{256} & \textbf{58.01} & \textbf{64.60} & \textbf{62.01} & \textbf{69.94} & \textbf{63.64} \\
512 & 57.81 & 63.97 & 62.08 & 68.49 & 63.09 \\
1024 & 55.54 & 62.44 & 61.12 & 67.79 & 61.72 \\
\bottomrule
\end{tabular}
\caption{mAP for cross-val split for Model: faster\_rcnn\_R\_50\_FPN\_3x}
\label{tab:faster_hyper_batch}
\end{table}

We run experiments with learning rates 1e-3, 1e-4 and 1e-5, with a fixed batch size of 2 images and batch size per image of 512 (which signifies the number of region proposals generated per image in the Region of Interest Pooling layer). The number of iterations trained for is 5,000. The results for the Faster R-CNN model are presented in Tab.~\ref{tab:faster_hyper_lr}. As observed, the best results are obtained for a learning rate of $1e-4$, so this parameter is fixed. Then, another hyperparameter search is performed to find the best batch size per image from 256, 512 and 1024. The results are presented in Table \ref{tab:faster_hyper_batch}. In this case, the best performance is obtained with a Batch size per image (ROI) of 256, so we fix it to this value. For RetinaNet, the best learning rate is $1e-3$ and the batch size per image 512.

%----------------SEGMENTATION
\section{Instance Segmentation}

\subsection{Pre-trained models: Qualitative and Quantitative Results}
\label{sec:app_mask_qualitative_quantitative}

\begin{figure}[b!]
    \centering
    \includegraphics[width=\linewidth]{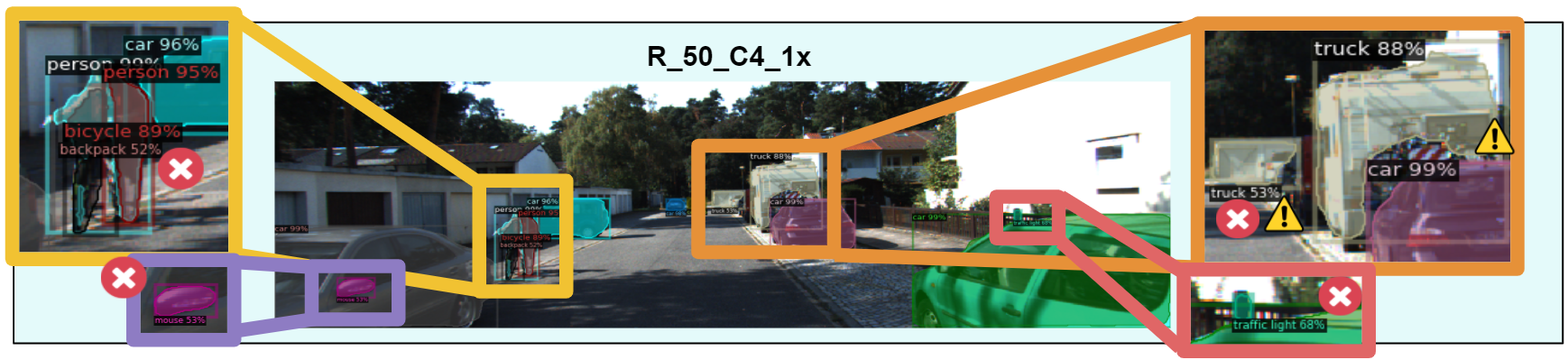}
    (a)
    \includegraphics[width=\linewidth]{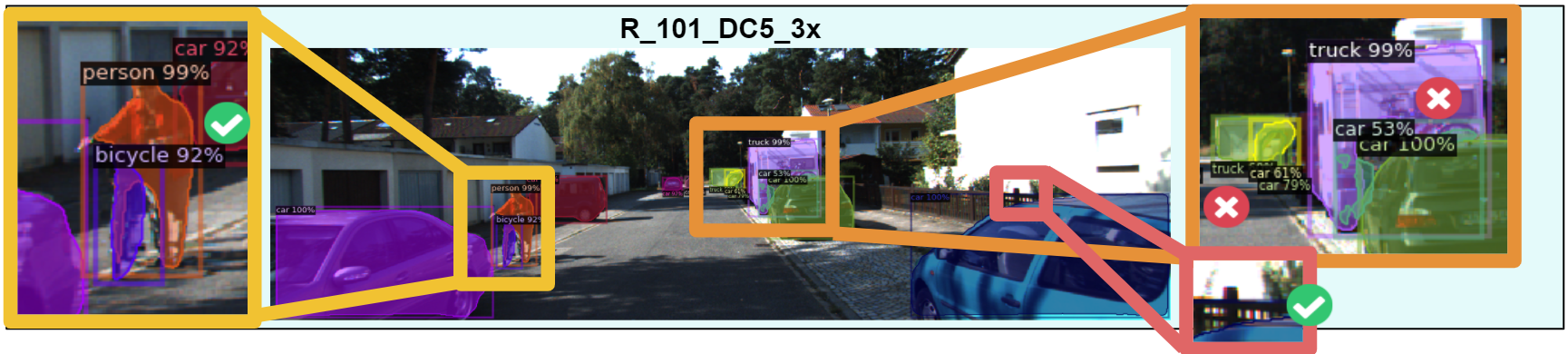}
    (b)
    \includegraphics[width=\linewidth]{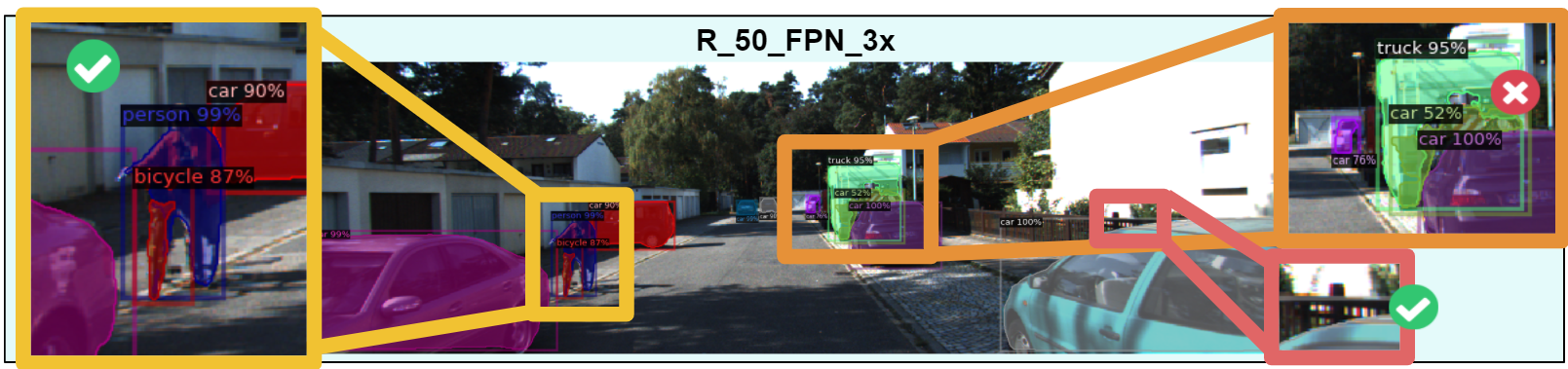}
    (c)
    \includegraphics[width=\linewidth]{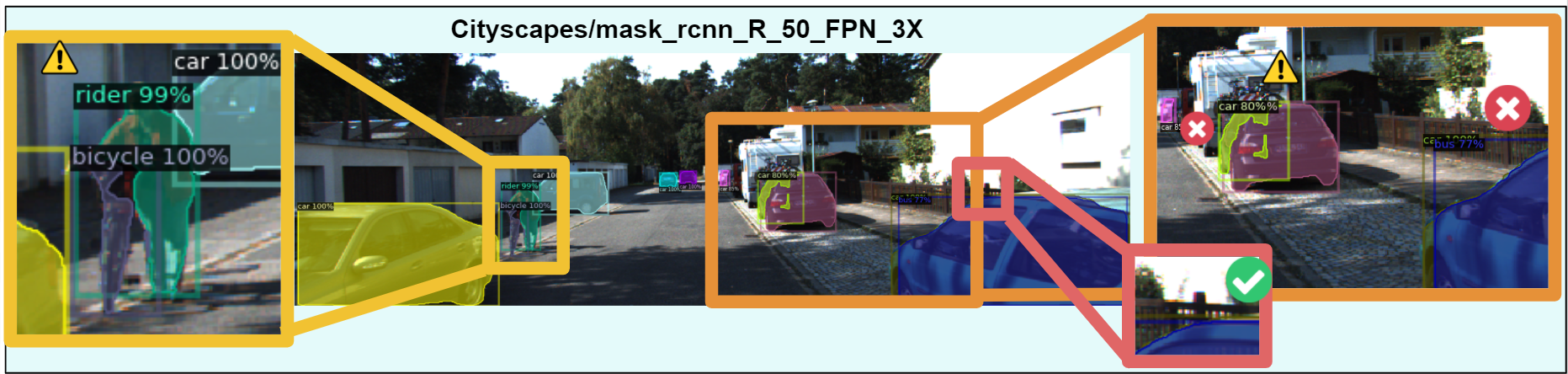}
    (d)
    \caption{Results from pre-trained Mask-RCNN models from Model Zoo: (a) R\_50\_C4\_1x, (b) R\_101\_DC5\_3x, (c) R\_50\_FPN\_3x and (d) R\_50\_FPN\_3x with Cityscapes}
    \label{fig:qualitative_mask}
\end{figure}

In Fig.~\ref{fig:qualitative_mask} we present the qualitative results for some Mask R-CNN model configurations. 
Using R\_50\_C4\_1x (Fig.~\ref{fig:qualitative_mask}a), we can observe multiple errors. The person walking with a bicycle on the left part of the image is detected as two different persons, one on each side of the bicycle. It fails to deal with the occlusion generated by the bike.  Additionally, it detects a false positive (FP) of a backpack. The same happens with the mirror of the car, that is detected as a mouse. On the right side of the image we can detect two additional FPs: a parking door at the end of the street is detected as a truck, and a part of the fence detected as a traffic light. Finally, we observe two miss detections on the right side of the image: there is a car not being detected at the end of the street, and the bicycle between the mobile home and car is also missed.

On the other hand, using R\_101\_DC5\_3x (Fig.~\ref{fig:qualitative_mask}b), the model deals correctly with the person detection and the bicycle on the left, it detects the parked car at the end of the street (right) and solves the FPs of the mouse (left) and traffic light (right). However, it detects a FP truck on the garage at the end of the street and the FP car behind the mobile home (right).

\begin{table*}[t!]
\centering
\begin{tabular}{cccccccc|cccc}
\toprule
 & \multicolumn{11}{c}{{Model: Mask\_RCNN\_}} \\ 
\cmidrule{2-11}
& \multicolumn{7}{c|}{{R-50-X}} & \multicolumn{4}{c}{{R-101-X}}  \\
Metric & C4-1x & DC5-1x & FPN-1x & C4-3x & DC5-3x & FPN-3x & Cityscapes & C4-3x & DC5-3x & FPN-3x  & X-FPN-3x\\ 
\midrule
AP   & 55.33 & 52.22 & 54.81 & 54.40 & 54.33 & \textbf{56.96} & 51.11 & 55.22 & 54.71 & 55.68 & 57.99\\
APs  & 36.01 & 34.79 & 39.70 & 37.55 & 36.76 & \textbf{40.94} & 36.33 & 37.19 & 37.05 & 38.70 & 41.48 \\
APm  & 68.76 & 64.22 & 65.06 & 68.06 & 66.33 & \textbf{67.91} & 60.93 & 68.03 & 66.68 & 67.30 & 68.72 \\
APl  & 64.96 & 60.21 & 62.83 & 61.48 & 60.98 & \textbf{64.11} & 64.76 & 62.78 & 62.49 & 61.09 & 65.92 \\
\bottomrule
\end{tabular}
\caption{Pre-trained Mask R-CNN Models \textbf{Bounding Box Detection} evaluated on the KITTI MOTS validation dataset}
\label{tab:mask_pre_trained_box}
\end{table*}
\begin{table*}[t!]
\centering
\begin{tabular}{cccccccc|cccc}
\toprule
 & \multicolumn{11}{c}{{Model: Mask\_RCNN\_}} \\ 
\cmidrule{2-11}
& \multicolumn{7}{c|}{{R-50-X}} & \multicolumn{4}{c}{{R-101-X}}  \\
Metric & C4-1x & DC5-1x & FPN-1x & C4-3x & DC5-3x & FPN-3x & Cityscapes & C4-3x & DC5-3x & FPN-3x & X-FPN-3x\\ 
\midrule
AP   & 41.82 & 40.32 & 43.76 & 42.72 & 42.57 & \textbf{45.90} & 41.45 & 42.61 & 43.17 & 44.35 & 46.20\\
APs  & 22.52 & 21.44 & 26.64 & 23.77 & 23.26 & \textbf{27.93} & 26.45 & 22.74 & 23.42 & 25.02 & 26.91 \\
APm  & 51.73 & 49.97 & 53.19 & 52.61 & 52.39 & \textbf{55.79} & 50.83 & 52.96 & 53.00 & 54.89 & 56.23 \\
APl  & 65.24 & 62.68 & 62.70 & 65.08 & 65.78 & \textbf{65.18} & 57.11 & 66.28 & 67.28 & 66.74 & 68.11 \\
\bottomrule
\end{tabular}
\caption{Pre-trained Mask R-CNN Models \textbf{Mask Detection} evaluated on the KITTI MOTS validation dataset}
\label{tab:mask_pre_trained_seg}
\end{table*}

The R\_50\_FPN\_3x model (Fig.~\ref{fig:qualitative_mask}c) correctly detects the pedestrian and bicycle on the left, the parked car at the end of the street (right) and does not detect the FPs of the  truck on the garage door (right) and the traffic light on the fence (right). However, there is a FP of a car behind the mobile home on the right.

Finally, the results obtained with model R\_50\_FPN\_3x trained with COCO and Cityscapes are presented in Fig.~\ref{fig:qualitative_mask}d. On the left side, we observe not an error per se, but a difference in the class definition, as Cityscapes defines those persons on a bicycle as riders. On the right, we can observe how the car just behind the mobile home is detected twice, which introduces a FP. The model also fails to detect the mobile home and introduces a double detection of the car on the bottom right corner of the image. In qualitative terms, we obtain the best results when using R\_50\_FPN\_3x model only training with COCO.

We summarize the results from these experiments in Tables \ref{tab:mask_pre_trained_box} and \ref{tab:mask_pre_trained_seg} for object detection and instance segmentation, respectively. 

\subsection{Training on different Datasets: Qualitative Results}
\label{sec:app_mask_train_datasets}
\begin{figure}[b!]
    \centering
    \includegraphics[width=\linewidth]{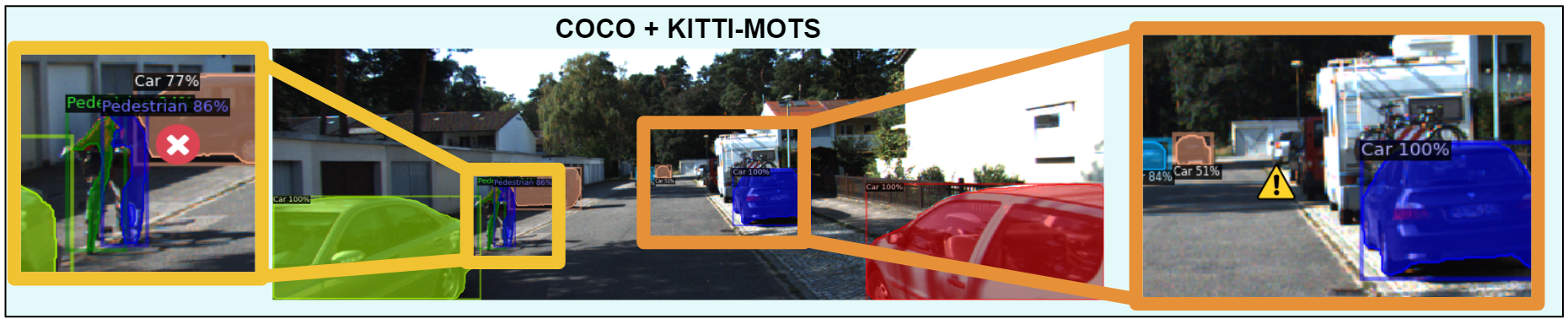}
    (a)
    \includegraphics[width=\linewidth]{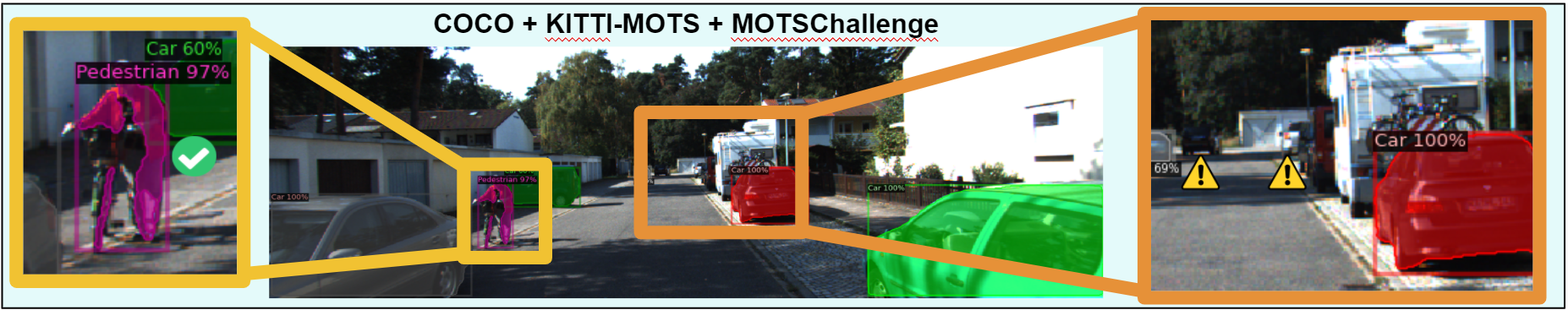}
    (b)
    \includegraphics[width=\linewidth]{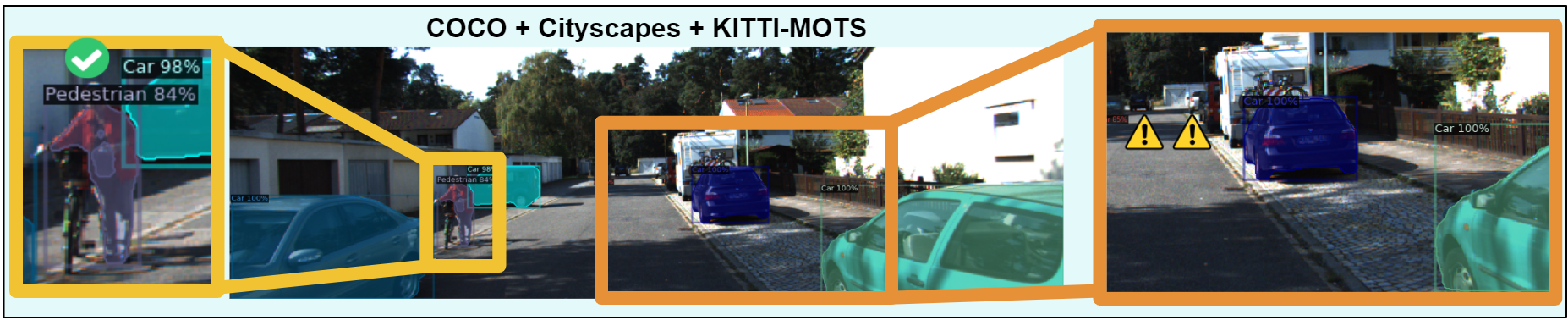}
    (c)
    \includegraphics[width=\linewidth]{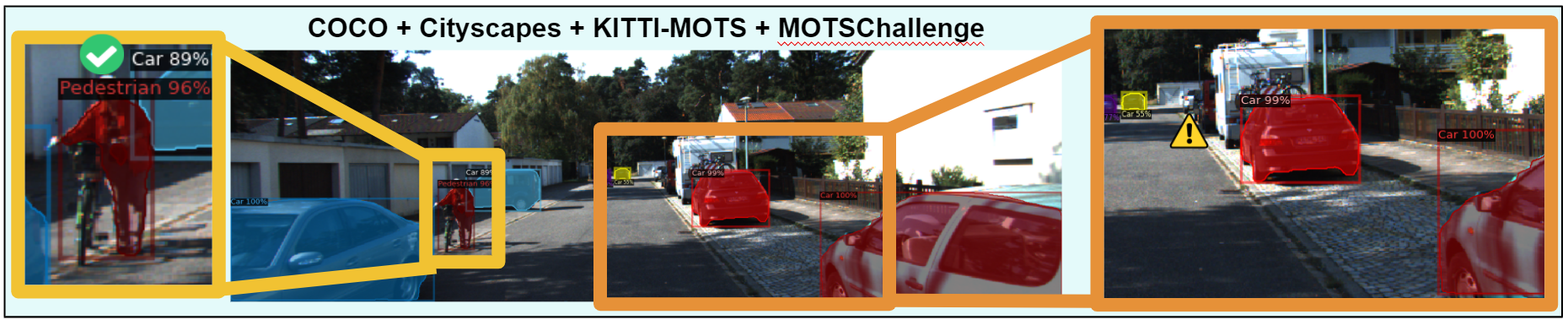}
    (d)
    \caption{Results from trained Mask-RCNN using: (a) COCO + KITTI-MOTS, (b) COCO + KITTI-MOTS + MOTSChallenge, (c) COCO + Cityscapes + KITTI-MOTS and (d) COCO + Cityscapes + KITTI-MOTS + MOTSChallenge}
    \label{fig:qualitative_mask_datasets}
\end{figure}

The qualitative results are presented in Fig.~\ref{fig:qualitative_mask_datasets}. As observed in Fig.~\ref{fig:qualitative_mask_datasets}a, when training with COCO and KITTI-MOTS, the model wrongly detects two pedestrians on the left part of the image, as it fails to deal with the occlusion caused by the bicycle. It also misses the car parked at the end of the street (right).

When including the MOTSChallenge dataset (Fig.~\ref{fig:qualitative_mask_datasets}b), which only has pedestrian labels, the double detection of the pedestrian with the bicycle on the left is solved. It seems that the precision regarding the pedestrian class has improved (but we cannot ensure anything until we check the quantitative results). However, the accuracy of the model regarding car detection decreases, and not only still misses the parked car at the end of the street (right), but also misses the car driving next to it.

On the other hand, when using COCO, Cityscapes and KITTI-MOTS for training (Fig.~\ref{fig:qualitative_mask_datasets}c), the behavior is the same than in the previous case: the model correctly detects the pedestrian on the left, but the car parked at the end of the street (right) and the other car driving next to it are missed.

Finally, when including MOTSChallenge dataset to the latter case (Fig.~\ref{fig:qualitative_mask_datasets}d), the confidence of detection on the pedestrian on the left increases, and the confidence of detections of some cars decreases. Surprisingly, even if the MOTSChallenge dataset only present labels for the pedestrian class, now the model is able to detect the car driving at the end of the road (right). However, we cannot get any conclusions from just a qualitative analysis of a single frame, so we study the results quantitatively.

\subsection{Hyperparameter Optimization}
\label{sec:app_mask_hyperparameter}
The first hyperparameter we optimize is the learning rate scheduler. We run experiments without using any scheduler and using One-Cycle policy \cite{one_cycle}. The results, presented in Table \ref{tab:iou_scheduler}, show that using One Cycle Scheduler improve the performance in 60\% less iterations, so it is the one we will use from now on.

Mask R-CNN uses a ResNet as backbone, which is divided in 5 stages: the first is a convolution and the following ones are each group of residual blocks. When fine tuning Mask R-CNN with Detectron2, by default 2 stages of ResNet are frozen and the rest retrained. However, this parameter can be changed to unfreeze and retrain more or less stages. The corresponding results are presented in Table ~\ref{tab:iou_frozen}. As observed, the best results are obtained with the default configuration, where two stages of layers are frozen and the rest are fine-tuned.

We also experiment with the background and foreground IoU thresholds for the anchors generated by the model. If the IoU threshold of the anchor lies between these 2 values, then the anchor can neither be classified as background or foreground and is ignored by the model. The results, presented in Table \ref{tab:iou_threshold}, show that the best results are obtained using the default configuration: 0.3 and 0.7 for the background and foreground thresholds, respectively.

Finally, we try different Non-Maximum Suppression (NMS) thresholds. This parameter sets the minimum overlap between bounding boxes needed so that they are merged. This way, the highly-overlapping bounding boxes of a same object are merged into a single one. The best results are obtained using a NMS threshold of 0.7, as shown in Table~\ref{tab:nms}.

\begin{table}[t!]
\centering
\begin{tabular}{cc|cccc|c}
\toprule
Scheduler & \# iters & 0 & 1 & 2 & 3 & Mean AP \\
\midrule
None & 5,000 & 54.09 & 47.53 & 49.21 & 34.92 & 46.44 \\
One Cycle & 2,000 & 52.79 & 49.76 & 49.30 & 35.36 & 46.80\\
\bottomrule
\end{tabular}
\caption{\textbf{Mask mAP} for cross-val split for Model: mask\_rcnn\_R\_50\_FPN\_3x with and without One Cycle Scheduler}
\label{tab:iou_scheduler}
\end{table}

\begin{table}[t!]
\centering
\begin{tabular}{c|cccc|c}
\toprule
frozen stages & 0 & 1 & 2 & 3 & Mean AP \\
\midrule
1 & 51.99 & 49.11 & 49.34 & 36.04 & 46.62 \\
2 & 52.78 & 49.76 & 49.30 & 35.36 & 46.80 \\
3 & 52.35 & 48.77 & 48.73 & 34.99 & 46.21 \\
4 & 51.53 & 46.78 & 47.22 & 36.14 & 45.42 \\
\bottomrule
\end{tabular}
\caption{\textbf{Mask mAP} for cross-val split for Model: mask\_rcnn\_R\_50\_FPN\_3x after freezing different number of stages}
\label{tab:iou_frozen}
\end{table}

\begin{table}[t!]
\centering
\begin{tabular}{c|cccc|c}
\toprule
bg and fg & 0 & 1 & 2 & 3 & Mean AP \\
\midrule
0.3 and 0.7 & 65.10 & 62.93 & 61.83 & 52.28 & 60.54 \\
0.5 and 0.5 & 65.19 & 62.52 & 62.43 & 46.64 & 59.19 \\
0.1 and 0.9 & 64.47 & 62.20 & 62.39 & 45.55 & 58.65 \\
\bottomrule
\end{tabular}
\caption{\textbf{Bounding Box mAP} for cross-val split for Model: mask\_rcnn\_R\_50\_FPN\_3x after applying background and foreground IoU thresholds}
\label{tab:iou_threshold}
\end{table}

\begin{table}[t!]
\centering
\begin{tabular}{c|cccc|c}
\toprule
NMS & 0 & 1 & 2 & 3 & Mean AP \\
\midrule
0.3 & 52.04 & 49.19 & 49.59 & 36.37 & 46.80 \\
0.5 & 52.22 & 49.19 & 49.64 & 36.48 & 46.88 \\
0.7 & 52.78 & 49.76 & 49.30 & 35.36 & 46.80 \\
0.9 & 52.88 & 49.98 & 50.45 & 36.03 & 47.34 \\
\bottomrule
\end{tabular}
\caption{\textbf{Mask mAP} for cross-val split for Model: mask\_rcnn\_R\_50\_FPN\_3x for different non-maximum suppression thresholds}
\label{tab:nms}
\end{table}

\end{appendices}

\newpage

\bibliographystyle{unsrt}
\bibliography{main}
\addtolength{\textheight}{-12cm}   % This command serves to balance the column lengths

\end{document}